\documentclass[11pt,a4paper]{article}
\usepackage[hyperref]{emnlp2020}
\usepackage{times}
\usepackage{latexsym}
\usepackage{xcolor}
\usepackage{booktabs}
\usepackage{graphicx}
\usepackage{tipa}
\usepackage{float}
\usepackage{amssymb,amsmath,amsthm,enumitem}
\usepackage{subcaption}
\usepackage{fdsymbol}

\usepackage{bm}

\usepackage{microtype}

\aclfinalcopy % Uncomment this line for the final submission
\def\aclpaperid{2493} %  Enter the acl Paper ID here

\newcommand{\xvec}{\mathbf{x}}
\newcommand{\tvec}{\mathbf{t}}

\newcommand{\hvec}{\mathbf{h}}

\newcommand{\cvec}{\mathbf{c}}
\newcommand{\svec}{\mathbf{s}}
\newcommand{\yvec}{\mathbf{y}}
\newcommand{\step}[1]{_{#1}}

\usepackage{rotating}
\usepackage{multirow}

\usepackage{tikz}
\usetikzlibrary{shapes.geometric}
\usetikzlibrary{patterns}
\usepackage{pgfplots}
\pgfplotsset{compat=1.15}

\DeclareMathOperator*{\argmax}{arg\,max}
\renewcommand{\arraystretch}{1.2}

\newcommand{\ba}{`}

\newcommand\blfootnote[1]{%
  \begingroup
  \renewcommand\thefootnote{}\footnote{\noindent #1}%
  \addtocounter{footnote}{-1}%
  \endgroup
}

\definecolor{burntred}{HTML}{E66101}
\definecolor{burntblue}{HTML}{00A2FF}

\title{OCR Post Correction for Endangered Language Texts}

\author{Shruti Rijhwani,\textsuperscript{$1$} Antonios Anastasopoulos,\textsuperscript{$2,\dagger$} Graham Neubig\textsuperscript{$1$}  \\
  \textsuperscript{$1$}Language Technologies Institute, Carnegie Mellon University\\
  \textsuperscript{$2$}Department of Computer Science, George Mason University\\
  \texttt{\{srijhwan,gneubig\}@cs.cmu.edu},\quad\texttt{antonis@gmu.edu}}

\date{}

\begin{document}

\maketitle
\begin{abstract}
There is little to no data available to build natural language processing models for most endangered languages. However, textual data in these languages often exists in formats that are not machine-readable, such as paper books and scanned images. In this work, we address the task of extracting text from these resources. We create a benchmark dataset of transcriptions for scanned books in three critically endangered languages and present a systematic analysis of how general-purpose OCR tools are not robust to the data-scarce setting of endangered languages. We develop an OCR post-correction method tailored to ease training in this data-scarce setting, reducing the recognition error rate by 34\% on average across the three languages.\blfootnote{$\dagger$: Work done at Carnegie Mellon University.}\footnote{Code and data are available at \url{https://shrutirij.github.io/ocr-el/}.}
\end{abstract}

\section{Introduction}
\begin{figure}[t]
\centering
    \begin{subfigure}[t]{\columnwidth}
    \centering
      \caption{Ainu (left) -- Japanese (right)}
      \vspace{-0.5em}
      \fbox{\includegraphics[width=0.9\columnwidth]{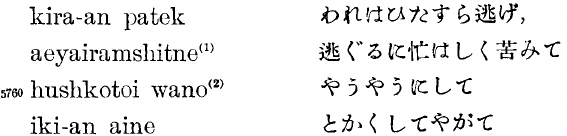}}
      \vspace{0.6em}
    \end{subfigure}
    \begin{subfigure}[t]{\columnwidth}
      \centering
      \caption{Griko (top) -- Italian (bottom)}
      \vspace{-0.5em}
      \frame{\includegraphics[width=0.93\columnwidth]{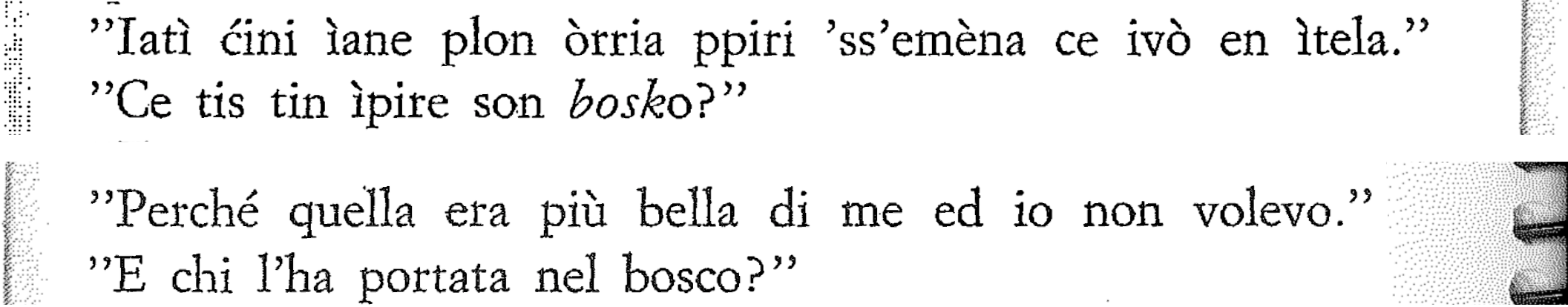}}
      \vspace{0.7em}
    \end{subfigure}
    \begin{subfigure}[t]{\columnwidth}
      \centering
      \caption{Yakkha (top) -- Nepali (middle) -- English (bottom)}
      \vspace{-0.5em}
      \fbox{\includegraphics[width=0.9\columnwidth]{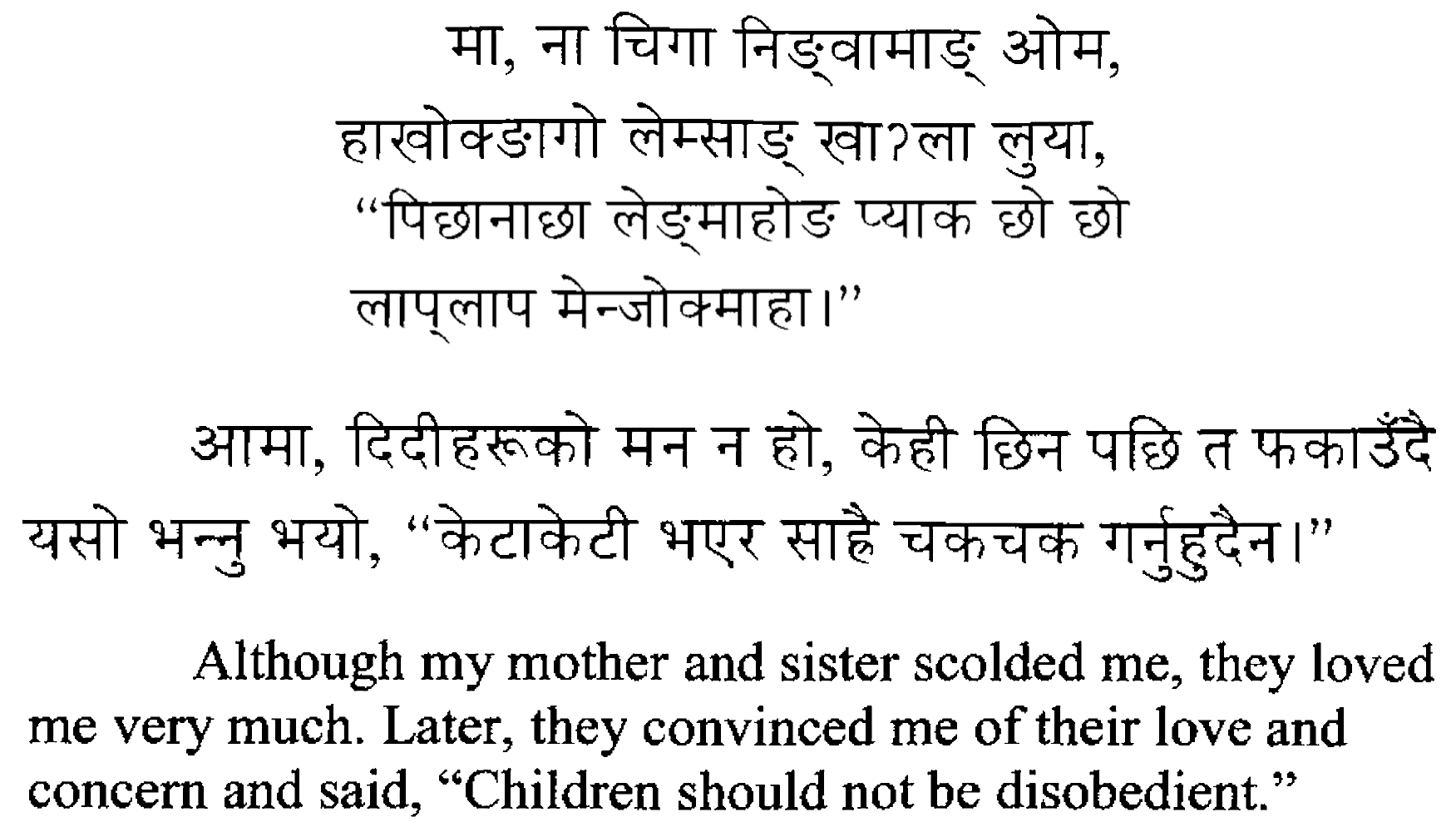}}
      \vspace{0.6em}
    \end{subfigure}
    \footnotesize{(d) Handwritten Shangaji -- typed English glosses}
    \begin{tabular}{|@{\ \ }c@{\ \ }|}
    \hline
     \begin{subfigure}[t]{0.9\columnwidth}
      \centering
      \includegraphics[width=0.9\columnwidth]{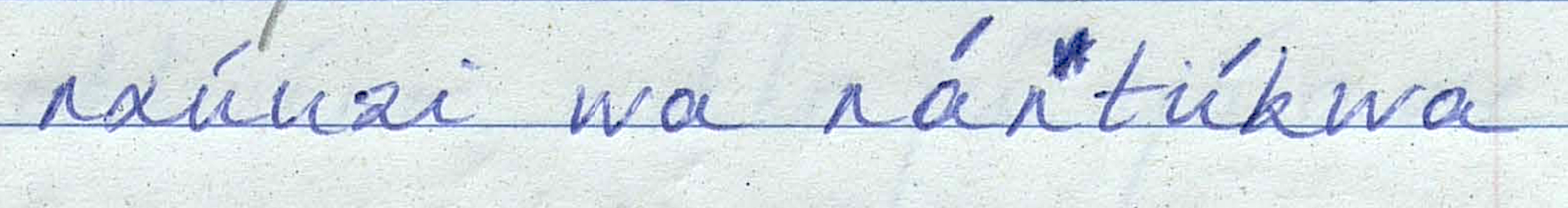}
      \includegraphics[width=0.5\columnwidth]{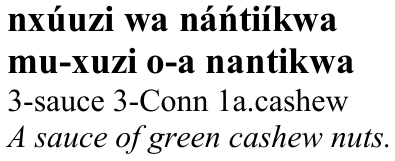}
    \end{subfigure}\\
    \hline
    \end{tabular}
    \caption{Examples of scanned documents in endangered languages accompanied by translations from the same scanned book (a, b, c) or linguistic archive (d).}
    \label{fig:dataset_example}
    \vspace{-1.2em}
\end{figure}

Natural language processing (NLP) systems exist for a small fraction of the world's over 6,000 living languages, the primary reason being the lack of resources required to train and evaluate models. Technological advances are concentrated on languages that have readily available data, and most other languages are left behind~\cite{joshi2020state}. This is particularly notable in the case of endangered languages, i.e., languages that are in danger of becoming extinct due to dwindling numbers of native speakers and the younger generations shifting to using other languages. For most endangered languages, finding \emph{any} data at all is challenging.

In many cases, natural language text in these languages does exist. However, it is locked away in formats that are not machine-readable --- paper books, scanned images, and unstructured web pages. These include books from local publishing houses within the communities that speak endangered languages, such as educational or cultural materials. Additionally, linguists documenting these languages also create data such as word lists and interlinear glosses, often in the form of handwritten notes. Examples from such scanned documents are shown in~\autoref{fig:dataset_example}. Digitizing the textual data from these sources will not only enable NLP for endangered languages but also aid linguistic documentation, preservation, and accessibility efforts.

In this work, we create a benchmark dataset and propose a suite of methods to extract data from these resources, focusing on scanned images of paper books containing endangered language text. Typically, this sort of digitization requires an optical character recognition (OCR) system. However, the large amounts of textual data and transcribed images needed to train state-of-the-art OCR models from scratch are unavailable in the endangered language setting. Instead, we focus on \emph{post-correcting} the output of an off-the-shelf OCR tool that can handle a variety of scripts. We show that targeted methods for post-correction can significantly improve performance on endangered languages.

Although OCR post-correction is relatively well-studied, most existing methods rely on considerable resources in the target language, including a substantial amount of textual data to train a language model~\cite{schnober-etal-2016-still,dong-smith-2018-multi,8978127} or to create synthetic data~\cite{krishna-etal-2018-upcycle}. While readily available for high-resource languages, these resources are severely limited in endangered languages, preventing the direct application of existing post-correction methods in our setting. 

As an alternative, we present a method that compounds on previous models for OCR post-correction, making three improvements tailored to the data-scarce setting. First, we use a \textbf{multi-source model} to incorporate information from the high-resource translations that commonly appear in endangered language books. These translations are usually in the \textit{lingua franca} of the region (e.g., \autoref{fig:dataset_example} (a,b,c)) or the documentary linguist's primary language (e.g., \autoref{fig:dataset_example} (d) from \citet{shangaji-elar}). Next, we introduce \textbf{structural biases} to ease learning from small amounts of data. Finally, we add \textbf{pretraining methods} to utilize the little unannotated data that exists in endangered languages.

\medskip
\noindent
We summarize our main contributions as follows:
\begin{itemize}[leftmargin=*, itemsep=3pt]
    \item A benchmark dataset for OCR post-correction on three critically endangered languages: Ainu, Griko, and Yakkha.
    \item A systematic analysis of a general-purpose OCR system, demonstrating that it is not robust to the data-scarce setting of endangered languages.
    \item An OCR post-correction method that adapts the standard neural encoder-decoder framework to the highly under-resourced endangered language setting, reducing both the character error rate and the word error rate by 34\% over a state-of-the-art general-purpose OCR system.
\end{itemize}

\section{Problem Setting}
\label{sec:setting}
In this section, we first define the task of OCR post-correction and introduce how we incorporate translations into the correction model. Next, we discuss the sources from which we obtain scanned documents containing endangered language texts.

\subsection{Formulation}
\label{sec:formulation}
\paragraph{Optical Character Recognition} OCR tools are trained to find the best transcription corresponding to the text in an image. The system typically consists of a recognition model that produces candidate text sequences conditioned on the input image and a language model that determines the probability of these sequences in the target language. We use a general-purpose OCR system (detailed in \autoref{sec:analysis}) to produce a \emph{first pass transcription} of the endangered language text in the image. Let this be a sequence of characters $\boldsymbol{x} = [x_1, \dots, x_N]$.

\paragraph{OCR post-correction} The goal of post-correction is to reduce recognition errors in the first pass transcription --- often caused by low quality scanning, physical deterioration of the paper book, or diverse layouts and typefaces~\cite{dong-smith-2018-multi}. The focus of our work is on using post-correction to counterbalance the lack of OCR training data in the target endangered languages. The correction model takes $\boldsymbol{x}$ as input and produces the \emph{final transcription} of the endangered language document, a sequence of characters $\boldsymbol{y} = [y_1, \dots , y_K]$. 
$$\boldsymbol{y} = \argmax_{\boldsymbol{y'}} p_\text{corr}(\boldsymbol{y'}|\boldsymbol{x})$$

\noindent
\textbf{Incorporating translations}\quad We use information from high-resource translations of the endangered language text. These translations are commonly found within the same paper book or linguistic archive (e.g., \autoref{fig:dataset_example}). We use an existing OCR system to obtain a transcription of the scanned translation, a sequence of characters $\boldsymbol{t} = [t_1, \dots, t_M]$. This is used to condition the model:
$$\boldsymbol{y} = \argmax_{\boldsymbol{y'}} p_\text{corr}(\boldsymbol{y'}|\boldsymbol{x}, \boldsymbol{t})$$

\subsection{Endangered Language Documents}
We explore online archives to determine how many scanned documents in endangered languages exist as potential sources for data extraction (as of this writing, October 2020).

The Internet Archive,\footnote{\url{https://archive.org/}} a general-purpose archive of web content, has scanned books labeled with the language of their content. We find 11,674 books labeled with languages classified as \ba\ba endangered'' by UNESCO.
Additionally, we find that endangered language linguistic archives contain thousands of documents in PDF format --- the Archive of the Indigenous Languages of Latin America (AILLA)\footnote{\url{https://ailla.utexas.org}} contains $\approx$10,000 such documents and the Endangered Languages Archive (ELAR)\footnote{\url{https://elar.soas.ac.uk/}} has $\approx$7,000. 

\medskip
\noindent
\textbf{How common are translations?} As described in the introduction, endangered language documents often contain a translation into another (usually high-resource) language. While it is difficult to estimate the number of documents with translations, multilingual documents represent the majority in the archives we examined; AILLA contains 4,383 PDFs with bilingual text and 1,246 PDFs with trilingual text, while ELAR contains $\approx$5,000 multilingual documents. The structure of translations in these documents is varied, from dictionaries and interlinear glosses to scanned multilingual books.

\section{Benchmark Dataset}
\label{sec:dataset}
From the sources described above, we select documents from three critically endangered languages\footnote{UNESCO defines critically endangered languages as those where the youngest speakers are grandparents and older, and they speak the language partially and infrequently.} for annotation --- Ainu, Griko, and Yakkha. These languages were chosen in an effort to create a geographically, typologically, and orthographically diverse benchmark. We focus this initial study on scanned images of printed books as opposed to handwritten notes, which are a relatively more challenging domain for OCR. 

We manually transcribed the text corresponding to the endangered language content. The text corresponding to the translations is not manually transcribed. We also aligned the endangered language text to the OCR output on the translations, per the formulation in \autoref{sec:formulation}. We describe the annotated documents below and example images from our dataset are in \autoref{fig:dataset_example} (a), (b), (c).

\smallskip
\textbf{Ainu} is a severely endangered indigenous language from northern Japan, typically considered a language isolate. In our dataset, we use a book of Ainu epic poetry (\textit{yukara}), with the ``Kutune Shirka" yukara~\cite{kindaichi1931ainu} in Ainu transcribed in Latin script.\footnote{Some transcriptions of Ainu also use the Katakana script. See \citet{howell1951classification} for a discussion on Ainu folklore.} Each page in the book has a two-column structure --- the left column has the Ainu text, and the right has its Japanese translation already aligned at the line-level, removing the need for manual alignment (see \autoref{fig:dataset_example} (a)). The book has 338 pages in total. Given the effort involved in annotation, we transcribe the Ainu text from 33 pages, totaling 816 lines.

\smallskip
\textbf{Griko} is an endangered Greek dialect spoken in southern Italy. The language uses a combination of the Latin alphabet and the Greek alphabet as its writing system. The document we use is a book of Griko folk tales compiled by \citet{stomeo1980racconti}. The book is structured such that in each fold of two pages, the left page has Griko text, and the right page has the corresponding translation in Italian. Of the 175 pages in the book, we annotate the Griko text from 33 pages and manually align it at the sentence-level to the Italian translation. This results in 807 annotated Griko sentences.

\smallskip
\textbf{Yakkha} is an endangered Sino-Tibetan language spoken in Nepal. It uses the Devanagari writing system. We use scanned images of three children's books, each of which has a story written in Yakkha along with its translation in Nepali and English~\cite{yakkha-elar}. We manually transcribe the Yakkha text from all three books. We also align the Yakkha text to both the Nepali and the English OCR at the sentence level with the help of an existing Yakkha dictionary~\cite{Schackow_2015}. In total, we have 159 annotated Yakkha sentences.

\section{OCR Systems: Promises and Pitfalls}
\label{sec:analysis}
As briefly alluded to in the introduction, training an OCR model for each endangered language is challenging, given the limited available data. 
Instead, we use the general-purpose OCR system from the Google Vision AI toolkit\footnote{\url{https://cloud.google.com/vision}} to get the first pass OCR transcription on our data.

The Google Vision OCR system~\cite{fujii2017sequence,ingle2019scalable} is highly performant and supports 60 major languages in 29 scripts. It can transcribe a wide range of higher resource languages with high accuracy, ideal for our proposed method of incorporating high-resource translations into the post-correction model. Moreover, it is particularly well-suited to our task because it provides script-specific OCR models in addition to language-specific ones. Per-script models are more robust to unknown languages because they are trained on data from multiple languages and can act as a general character recognizer without relying on a single language's model. Since most endangered languages adopt standard scripts (often from the region's dominant language) as their writing systems, the per-script recognition models can provide a stable starting point for post-correction.

The metrics we use for evaluating performance are character error rate (CER) and word error rate (WER), representing the ratio of erroneous characters or words in the OCR prediction to the total number in the annotated transcription. More details are in \autoref{sec:experiments}. The CER and WER using the Google Vision OCR on our dataset are in \autoref{tab:google_metrics}.

\subsection{OCR Performance}
\begin{table}[tb]
    \centering
    \small
    \begin{tabular}{lcrr}
    \toprule
    Language && CER & WER \\
    \midrule
    Ainu && 1.34 & 6.27 \\
    Griko && 3.27 & 15.63 \\
    Yakkha && 8.90 & 31.64 \\
    \bottomrule
    \end{tabular}
    \caption{Character error rate and word error rate with the Google Vision OCR system on our dataset.}
    \label{tab:google_metrics}
\end{table}
Across the three languages, the error rates indicate that we have a first pass transcription that is of reasonable quality, giving our post-correction method a reliable starting point. We note the particularly low CER for the Ainu data, reflecting previous work that has evaluated the Google Vision system to have strong performance on typed Latin script documents \cite{fujii2017sequence}. However, there remains considerable room for improvement in both CER and WER for all three languages.

Next, we look at the edit distance between the predicted and the gold transcriptions, in terms of insertion, deletion, and replacement operations. Replacement accounts for over 84\% of the errors in the Griko and Ainu datasets, and 55\% overall. This pattern is expected in the OCR task, as the recognition model uses the image to make predictions and is more likely to confuse a character's shape for another than to hallucinate or erase pixels. However, we observe that the errors in the Yakkha dataset do not follow this pattern. Instead, 87\% of the errors for Yakkha occur because of deleted characters.

\subsection{Types of Errors}

\begin{figure}[t]
    \centering
    \small
    \begin{tabular}{ccc}
        \frame{\includegraphics[width=0.4\columnwidth]{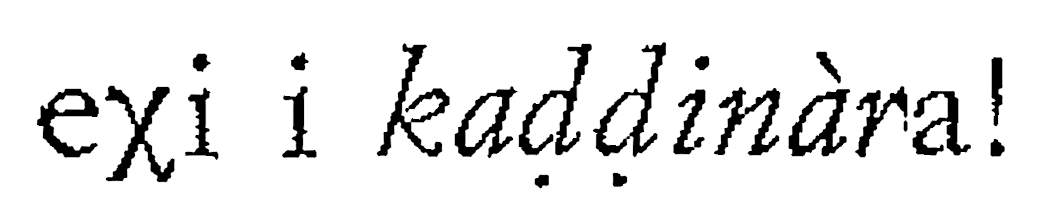}} & \raisebox{0.7em}{$\xrightarrow{\mathrm{OCR}}$} & \raisebox{0.7em}{\large e\textcolor{burntred}{\textbf{x}}i i ka\textcolor{burntred}{\textbf{dd}}in\`ara} \\[.1cm]
        \frame{\includegraphics[width=0.3\columnwidth]{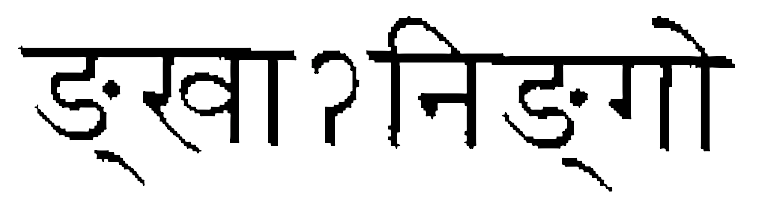}} &
        \raisebox{0.6em}{$\xrightarrow{\mathrm{OCR}}$} &
        \includegraphics[width=0.3\columnwidth]{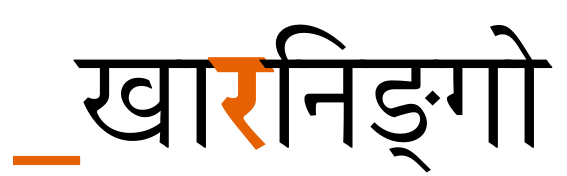} \\
    \end{tabular}
    \caption{Examples of errors in Griko (top) and Yakkha (bottom) when using the Google Vision OCR.}
    \label{fig:ocr_errors}
\end{figure}

To better understand the challenges posed by the endangered language setting, we manually inspect all the errors made by the OCR system. While some errors are commonly seen in the OCR task, such as misidentified punctuation or incorrect word boundaries, 85\% of the total errors occur due to specific characteristics of endangered languages that general-purpose OCR systems do not account for. Broadly, they can be categorized into two types, examples of which are shown in \autoref{fig:ocr_errors}:
\begin{itemize}[leftmargin=*, itemsep=2pt, topsep=6pt]
    \item\textbf{Mixed scripts}\quad The existing scripts that most endangered languages adopt as writing systems are often not ideal for comprehensively representing the language. For example, the Devanagari script does not have a grapheme for the glottal stop --- as a solution, printed texts in the Yakkha language use the IPA symbol \ba\textipa{\textglotstop}'~\cite{Schackow_2015}. Similarly, both Greek and Latin characters are used to write Griko. The Google Vision OCR is trained to detect script at the line-level and is not equipped to handle multiple scripts within a single word. As seen in \autoref{fig:ocr_errors}, the system does not recognize the Greek character $\bm{\chi}$ in Griko and the IPA symbol \textbf{\textipa{\textglotstop}} in Yakkha. Mixed scripts cause 11\% of the OCR errors.
    \item\textbf{Uncommon characters and diacritics}\quad Endangered languages often use graphemes and diacritics that are part of the standard script but are not commonly seen in high-resource languages. Since these are likely rare in the OCR system's training data, they are frequently misidentified, accounting for 74\% of the errors. In \autoref{fig:ocr_errors}, we see that the OCR system substitutes the uncommon diacritic \textbf{\d{d}} in Griko. The system also deletes the Yakkha character {\raisebox{-2.65pt}{\includegraphics[height=9.5pt]{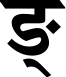}}}, which is a \ba half form' alphabet that is infrequent in several other Devanagari script languages (such as Hindi).
    
\end{itemize}
\section{OCR Post-Correction Model}
\label{sec:model}
In this section, we describe our proposed OCR post-correction model. The base architecture of the model is a multi-source sequence-to-sequence framework~\cite{zoph-knight-2016-multi,libovicky-helcl-2017-attention} that uses an LSTM encoder-decoder model with attention~\cite{bahdanau2015neural}. We propose improvements to training and modeling for the multi-source architecture, specifically tailored to ease learning in data-scarce settings.

\subsection{Multi-source Architecture}
\label{sec:base}
{%
\setlength{\fboxsep}{0pt}%
\setlength{\fboxrule}{1pt}%
}%
\begin{figure}[tb]
\tikzset{seq/.style={draw=none,fill=gray!20}}
\tikzset{layer/.style={->,thick}}
\tikzset{label/.style={anchor=west,font={\footnotesize}}}
\tikzset{seqlabel/.style={font={\small}}}
\newcommand{\encoder}[3]{
\draw[seq] (-1.25,-0.25) rectangle (1.25,0.25);
\node[seqlabel] at (0,0) 
%{$\xmat$};
{$#3\step{1}#1 \ldots #3\step{#2}#1$};
\draw[layer] (0,0.3) -- (0,0.7);
\node[seqlabel] at (0,0.5) [label] {encoder};
\draw[seq] (-1.25,0.75) rectangle (1.25,1.25);
\node[seqlabel] at (0,1) %{$\hmat$}; 
{$\hvec\step{1}#1 \ldots \hvec\step{#2}#1$};
}
\newcommand{\ocrencoder}[5]{
\node[inner sep=0pt] at (0,-1.7)
    {\small #5};
\node[inner sep=0pt] at (0,-1.2)
    {\setlength{\fboxsep}{.005\textwidth}%
    \fbox{\includegraphics[width=.145\textwidth]{#4}}};
\draw[layer] (0,-0.9) -- (0,-0.3);
\node[seqlabel] at (0,-0.6) [label] {\textsc{ocr}};
\draw[seq] (-1.25,-0.25) rectangle (1.25,0.25);
\node[seqlabel] at (0,0) 
%{$\xmat$};
{$#3\step{1} \ldots #3\step{#2}$};
\draw[layer] (0,0.3) -- (0,0.9);
\node[seqlabel] at (0,0.6) [label] {encoder};
\draw[seq] (-1.25,0.95) rectangle (1.25,1.45);
\node[seqlabel] at (0,1.2) %{$\hmat$}; 
{$\hvec\step{1}#1 \ldots \hvec\step{#2}#1$};
}
\newcommand{\decoder}[1]{
\draw[seq] (-1,2.05) rectangle (1,2.55);
\node[seqlabel] at (0,2.3) %{$\cmat#1$}; 
{$\cvec\step{1}#1 \ldots \cvec\step{K#1}#1$};
\draw[layer] (0,2.6) -- (0,3.2);
\node[seqlabel] at (0,2.9) [label] {decoder};
\draw[seq] (-1,3.25) rectangle (1,3.75);
\node[seqlabel] at (0,3.5) %{$\smat#1$}; 
{$\svec\step{1}#1 \ldots \svec\step{K#1}#1$};
\draw[layer] (0,3.8) -- (0,4.4);
\node[seqlabel] at (0,4.1) [label] {softmax};
\draw[seq] (-1,4.45) rectangle (1,5.05);
\node[seqlabel] at (0,4.75) %{$P(\ymat#1)$}; 
{$P(\yvec\step{1}#1 \ldots \yvec\step{K#1}#1)$};
}
\begin{center}
\resizebox{0.7\hsize}{!}{
\begin{tabular}{c}
\begin{tikzpicture}
\begin{scope}[xshift=-1.4cm]
% \encoder{^x}{N}{\xvec}
\ocrencoder{^x}{N}{\xvec}{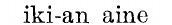}{Ainu}
\end{scope}
\begin{scope}[xshift=1.4cm]
% \encoder{^t}{M}{\tvec}
\ocrencoder{^t}{M}{\tvec}{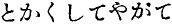}{Japanese}
\end{scope}
\draw[layer] (-1.4,1.5) -- (0,2.0);
\draw[layer] (1.4,1.5) -- (0,2.0);
\node at (-1,1.85) [label,anchor=east] {attention};
\node at (1,1.85) [label] {attention};
\decoder{}
\end{tikzpicture}
\end{tabular}%
}
\end{center}
\caption{The proposed multi-source architecture with the encoder for an endangered language segment (left) and an encoder for the translated segment (right). The input to the encoders is the first pass OCR over the scanned images of each segment. For example, the OCR on the scanned images of some Ainu text (left) and its Japanese translation (right).}
\label{fig:multisourcemodels}
\end{figure}
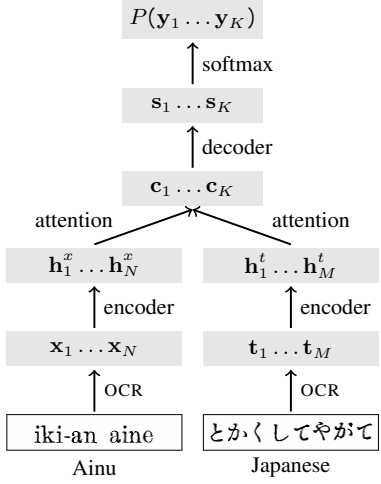
Our post-correction formulation takes as input the first pass OCR of the endangered language segment $\boldsymbol{x}$ and the OCR of the translated segment $\boldsymbol{t}$, to predict an error-free endangered language text $\boldsymbol{y}$. The model architecture is shown in \autoref{fig:multisourcemodels}.

The model consists of two encoders --- one that encodes $\boldsymbol{x}$ and one that encodes $\boldsymbol{t}$. Each encoder is a character-level bidirectional LSTM~\cite{hochreiter1997long} and transforms the input sequence of characters to a sequence of hidden state vectors: $\mathbf{h}^x$ for the endangered language text and $\mathbf{h}^t$ for the translation.

The model uses an attention mechanism during the decoding process to use information from the encoder hidden states. We compute the attention weights over each of the two encoders independently. At the decoding time step $k$:
\begin{align}
e^x_{k,i}=\mathbf{v}^x \tanh\left(\mathbf{W}_1^x \mathbf{s}_{k-1} + \mathbf{W}_2^x \mathbf{h}^x_i\right)
\label{eq:attn}
\end{align}
$$\boldsymbol{\alpha}_k^x = \mathrm{softmax}\left(\mathbf{e}_k^x\right)$$
$$\mathbf{c}^x_k = \left[\Sigma_i \alpha^x_{k,i} \mathbf{h}^x_i\right]$$
\noindent
where $\mathbf{s}_{k-1}$ is the decoder state of the previous time step and $\mathbf{v}^x$, $\mathbf{W}_1^x$ and $\mathbf{W}_2^x$ are trainable parameters. The encoder hidden states $\mathbf{h}^x$ are weighted by the attention distribution $\boldsymbol{\alpha}^x_k$ to produce the context vector $\mathbf{c}^x_k$. We follow a similar procedure for the second encoder to produce $\mathbf{c}^t_k$.
We concatenate the context vectors to combine attention from both sources~\cite{zoph-knight-2016-multi}:
$$\mathbf{c}_k=\left[\mathbf{c}_k^x;\mathbf{c}_k^t\right]$$
$\mathbf{c}_k$ is used by the decoder LSTM to compute the next hidden state $\mathbf{s}_k$ and subsequently, the probability distribution for predicting the next character $\mathbf{y}_k$ of the target sequence $\boldsymbol{y}$:
\begin{align} 
\mathbf{s}_k &= \mathrm{lstm}\left(\mathbf{s}_{k-1}, \mathbf{c}_k, \mathbf{y}_{k-1}\right)\\
P\left(\mathbf{y}_k\right) &= \mathrm{softmax}\left(\mathbf{W}\mathbf{s}_k + \mathbf{b}\right)
\label{eq:decoder}
\end{align}

\paragraph{Training and Inference} The model is trained for each language with the cross-entropy loss ($\mathcal{L}_\mathrm{ce}$) on the small amount of transcribed data we have. Beam search is used at inference time.

\subsection{Model and Training Improvements}
\label{sec:recipe1}

With the minimal annotated data we have, it is challenging for the neural network to learn a good distribution over the target characters. We propose a set of adaptations to the base architecture that improves the post-correction performance without additional annotation. The adaptations are based on characteristics of the OCR task itself and the performance of the upstream OCR tool (\autoref{sec:analysis}).

\paragraph{Diagonal attention loss} As seen in \autoref{sec:analysis}, substitution errors are more frequent in the OCR task than insertions or deletions; consequently, we expect the source and target to have similar lengths. Moreover, post-correction is a monotonic sequence-to-sequence task, and reordering rarely occurs~\cite{schnober-etal-2016-still}. 
Hence, we expect attention weights to be higher at characters close to the diagonal for the endangered language encoder.

We modify the model such that all the elements in the attention vector that are not within $j$ steps (we use $j=3$) of the current time step $k$ are added to the training loss, thereby encouraging elements away from the diagonal to have lower values. The diagonal loss summed over all time steps for a training instance, where $N$ is the length of $\boldsymbol{x}$, is:
$$\mathcal{L}_\mathrm{diag} = \sum_k \left(\sum_{i=1}^{k-j} \alpha^x_{k,i} + \sum_{i=k+j}^N \alpha^x_{k,i}\right)$$

\paragraph{Copy mechanism} \autoref{tab:google_metrics} indicates that the first pass OCR predicts a majority of the characters accurately. In this scenario, enabling the model to directly copy characters from the first pass OCR rather than generate a character at each time step might lead to better performance~\cite{gu-etal-2016-incorporating}.

We incorporate the copy mechanism proposed in~\citet{see-etal-2017-get}. The mechanism computes a \ba\ba generation probability'' at each time step $k$, which is used to choose between \emph{generating} a character (\autoref{eq:decoder}) or \emph{copying} a character from the source text by sampling from the attention distribution $\boldsymbol{\alpha}_k^x$.

\paragraph{Coverage} Given the monotonicity of the post-correction task, the model should not attend to the same character repeatedly. However, repetition is a common problem for neural encoder-decoder models~\cite{mi-etal-2016-coverage,tu-etal-2016-modeling}. To account for this problem, we adapt the coverage mechanism from~\citet{see-etal-2017-get}, which keeps track of the attention distribution over past time steps in a coverage vector. For time step $k$, the coverage vector would be $\mathbf{g}_k = \sum_{k'=0}^{k-1} \boldsymbol{\alpha}^x_{k'}$. 

$\mathbf{g}_k$ is used as an extra input to the attention mechanism, ensuring that future attention decisions take the weights from previous time steps into account. \autoref{eq:attn}, with learnable parameter $\mathbf{w}_g$, becomes:
$$e^x_{k,i}=\mathbf{v}^x \tanh\left(\mathbf{W}_1^x \mathbf{s}_{k-1} + \mathbf{W}_2^x \mathbf{h}^x_i + \mathbf{w}_g g_{k,i}\right)$$
$\mathbf{g}_k$ is also used to penalize attending to the same locations repeatedly with a coverage loss. The coverage loss summed over all time steps $k$ is:
$$\mathcal{L}_\mathrm{cov} = \sum_k \sum_{i=1}^n \min\left(\alpha_{k,i}^x, g_{k,i}\right)$$
Therefore, with our model adaptations, the loss for a single training instance:
\begin{align}
\mathcal{L} = \mathcal{L}_\mathrm{ce} + \mathcal{L}_\mathrm{diag} + \mathcal{L}_\mathrm{cov}
\label{eq:loss}
\end{align}

\subsection{Utilizing Untranscribed Data}
\label{sec:recipe2}

As discussed in \autoref{sec:dataset}, given the effort involved, we transcribe only a subset of the pages in each scanned book.
Nonetheless, we leverage the remaining unannotated pages for pretraining our model. We use the upstream OCR tool to get a first pass transcription on all the unannotated pages.

We then create \ba\ba pseudo-target'' transcriptions for the endangered language text as described below:
\begin{itemize}
    \item \textbf{Denoising rules}\quad Using a small fraction of the available annotated pages, we compute the edit distance operations between the first pass OCR and the gold transcription. The operations of each type (insertion, deletion, and replacement) are counted for each character and divided by the number of times that character appears in the first pass OCR. This forms a probability of how often the operation should be applied to that specific character.
    
    We use these probabilities to form rules, such as $p(\text{replace d with \d{d}})\!=\!0.4$ for Griko and $p(\text{replace ? with \textipa{\textglotstop}})\!=\!0.7$ for Yakkha. These rules are applied to remove errors from, or \ba\ba denoise'', the first pass OCR transcription.
    \item \textbf{Sentence alignment}\quad We use Yet Another Sentence Aligner~\cite{yasa-1336} for unsupervised alignment of the endangered language and translation on the unannotated pages. 
\end{itemize}
Given the aligned first pass OCR for the endangered language text and the translation along with the pseudo-target text, $\boldsymbol{x}$, $\boldsymbol{t}$ and $\boldsymbol{\hat{y}}$ respectively, the pretraining steps, in order, are:

\begin{itemize}
    \item \textbf{Pretraining the encoders}\quad We pretrain both the forward and backward LSTMs of each encoder with a character-level language model objective: the endangered language encoder on $\boldsymbol{x}$ and the translation encoder on $\boldsymbol{t}$.
    \item \textbf{Pretraining the decoder}\quad The decoder is pretrained on the pseudo-target $\boldsymbol{\hat{y}}$ with a character-level language model objective.
    \item \textbf{Pretraining the seq-to-seq model}\quad The model is pretrained with $\boldsymbol{x}$ and $\boldsymbol{t}$ as the sources and the pseudo-target $\boldsymbol{\hat{y}}$ as the target transcription, using the post-correction loss function~$\mathcal{L}$ as defined in \autoref{eq:loss}.
\end{itemize}

\section{Experiments}
\label{sec:experiments}
\begin{table*}[tb]
    \centering
    \small
    \begin{tabular}{l|r@{\ \ }rr@{\ \ }rr@{\ \ }r|r@{\ \ }rr@{\ \ }rr@{\ \ }r}
    \toprule
        & \multicolumn{6}{c|}{Character Error Rate} &\multicolumn{6}{c}{Word Error Rate} \\
        & \multicolumn{2}{c}{Ainu} & \multicolumn{2}{c}{Griko} & \multicolumn{2}{c|}{Yakkha} & \multicolumn{2}{c}{Ainu} & \multicolumn{2}{c}{Griko} & \multicolumn{2}{c}{Yakkha} \\[-0.3em]
        Model & \small Multi & \small Single & \small Multi & \small Single & \small Multi & \small Single & \small Multi & \small Single & \small Multi & \small Single & \small Multi & \small Single\\
        \midrule
        \textsc{Fp-Ocr} & \multicolumn{1}{c}{--} & $1.34$ & \multicolumn{1}{c}{--} & $3.27$ & \multicolumn{1}{c}{--} & $8.90$ & \multicolumn{1}{c}{--} & $6.27$ & \multicolumn{1}{c}{--} & $15.63$ & \multicolumn{1}{c}{--} & $31.64$ \\
        \textsc{Base} & $1.56$ & $1.41$ & $6.78$ & $5.95$ & $70.39$ & $71.71$ & $8.56$ & $7.88$ & $15.13$ & $13.67$ & $98.15$ & $99.10$  \\
        \textsc{Copy} & $2.04$ & $1.99$ & $2.54$ & $2.28$ & $14.77$ & $12.30$ & $9.48$ & $8.57$ & $9.33$ & $8.90$ & $30.36$ & $27.81$ \\
        \textsc{Ours} & $0.92$ & $\boldsymbol{0.80}$ & $\boldsymbol{1.66}$ & $1.70$ & $\boldsymbol{7.75}$ & $8.44$ & $5.75$ & $\boldsymbol{5.19}$ & $\boldsymbol{7.46}$ & $7.51$ & $\boldsymbol{20.95}$ & $21.33$ \\
    \bottomrule
    \end{tabular}
    \caption{Our method improves performance over all baselines (10-fold cross-validation averaged over five randomly seeded runs). We present multi- and single-source variants and \textbf{highlight} the best model for each language.}
    \label{tab:cer}
\end{table*}

This section discusses our experimental setup and the post-correction performance on the three endangered languages on our dataset.

\subsection{Experimental Setup}
\smallskip
\paragraph{Data Splits}
We perform 10-fold cross-validation for all experimental settings because of the small size of the datasets. For each language, we divide the transcribed data into 11 segments --- we use one segment for creating the \emph{denoising rules} described in the previous section and the remaining ten as the folds for cross-validation. In each cross-validation fold, eight segments are used for training, one for validation and one for testing.

We divide the dataset at the page-level for the Ainu and Griko documents, resulting in 11 segments of three pages each. For the Yakkha documents, we divide at the paragraph-level, due to the small size of the dataset. We have 33 paragraphs across the three books in our dataset, resulting in 11 segments that contain three paragraphs each. The multi-source results for Yakkha reported in \autoref{tab:cer} use the English translations. Results with Nepali are similar and are included in \autoref{sec:appendix}.

\paragraph{Metrics}
We use two metrics for evaluating our systems: character error rate (CER) and word error rate (WER). Both metrics are based on edit distance and are standard for evaluating OCR and OCR post-correction~\cite{berg-kirkpatrick-etal-2013-unsupervised,schulz-kuhn-2017-multi}. 
CER is the edit distance between the predicted and the gold transcriptions of the document, divided by the total number of characters in the gold transcription. WER is similar but is calculated at the word level.

\paragraph{Methods}
In our experiments, we compare the performance of our proposed method with the first pass OCR and with two systems from recent work in OCR post-correction. All the post-correction methods have two variants -- the single-source model with only the endangered language encoder and the multi-source model that additionally uses the high-resource translation encoder.
\begin{itemize}
    \item \textsc{Fp-Ocr}: The first pass transcription obtained from the Google Vision OCR system.
    \item \textsc{Base}: This system is the base sequence-to-sequence architecture described in \autoref{sec:base}. Both the single-source and multi-source variants of this system are used for English OCR post-correction in~\citet{dong-smith-2018-multi}. 
    \item \textsc{Copy}: This system is the base architecture with a copy mechanism as described in \autoref{sec:recipe1}. The single-source variant of this model is used for OCR post-correction on Romanized Sanskrit in \citet{krishna-etal-2018-upcycle}.\footnote{Although~\citet{krishna-etal-2018-upcycle} use BPE tokenization, preliminary experiments showed that character-level models result in much better performance on our dataset, likely due to the limited data available for training the BPE model.}
    \item \textsc{Ours}: The model with all the adaptations proposed in \autoref{sec:recipe1} and \autoref{sec:recipe2}.
\end{itemize}

\paragraph{Implementation} The post-correction models are implemented using the DyNet neural network toolkit~\cite{dynet}, and all reported results are the average of five training runs with different random seeds. We assume knowledge of the entire alphabet of the endangered language for all the methods, which is straightforward to obtain for most languages. The decoder's vocabulary contains all these characters, irrespective of their presence in the training data, with corresponding randomly-initialized character embeddings.

\subsection{Main Results}
\label{sec:results}
\renewcommand{\arraystretch}{1.0}
\begin{figure*}[tb]
    \centering
    \small
    \begin{tabular}{lcc}
    & \multicolumn{2}{c}{Errors \textit{fixed} by post-correction}\\[.1cm]
        & (a) Griko & (b) Yakkha  \\
        \raisebox{0.7em}{[Image]} & \frame{\includegraphics[width=0.4\columnwidth]{images/errors1a.pdf}} & \frame{\includegraphics[width=0.3\columnwidth]{images/errors2a.pdf}} \\
        & \multicolumn{2}{c}{$\big\downarrow$ \hspace{3cm} $\big\downarrow$}\\
        \raisebox{0.35em}{[First pass OCR]} & \raisebox{0.3em}{\large e\textcolor{burntred}{\textbf{x}}i i ka\textcolor{burntred}{\textbf{dd}}in\`ara} &
        \includegraphics[width=0.25\columnwidth]{images/error2b.pdf} \\
        & \multicolumn{2}{c}{$\big\downarrow$ \hspace{3cm} $\big\downarrow$}\\
        \raisebox{0.35em}{[Post-corrected]} & \raisebox{0.35em}{\large e\textcolor{burntblue}{$\bm{\chi}$}i i ka\textbf{\textcolor{burntblue}{\d{d}\d{d}}}in\`ara} &
        \includegraphics[width=0.28\columnwidth]{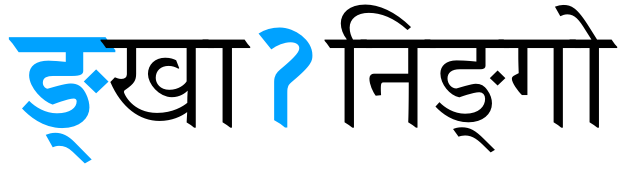} \\
    \end{tabular}
    \qquad
    \begin{tabular}{cc}
        \multicolumn{2}{c}{Errors \textit{introduced} by post-correction}\\[.1cm]
        (c) Griko & (d) Yakkha  \\
        \frame{\includegraphics[width=0.25\columnwidth]{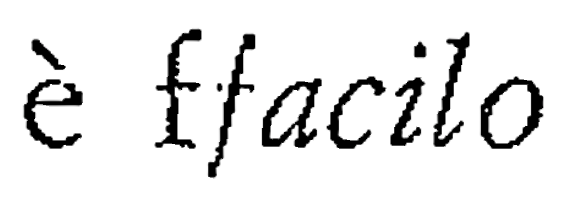}} & \frame{\includegraphics[width=0.35\columnwidth]{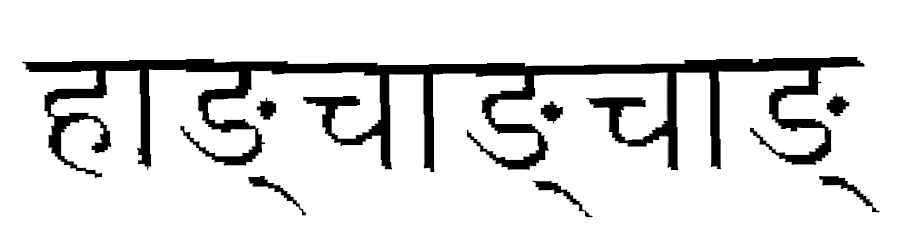}} \\
        \multicolumn{2}{c}{$\big\downarrow$ \hspace{2.5cm} $\big\downarrow$}\\
        \raisebox{0.35 em}{\large{\`{e} ffacilo}} &
        \includegraphics[width=0.3\columnwidth]{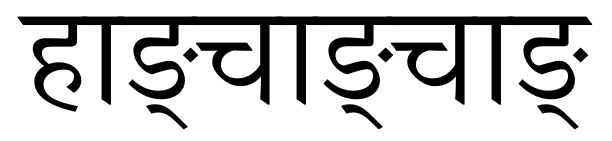} \\
        \multicolumn{2}{c}{$\big\downarrow$ \hspace{2.5cm} $\big\downarrow$}\\
        \raisebox{0.35 em}{\large{\`{e} ffa\textcolor{burntred}{\textbf{\'{c}}}ilo}} &
        \includegraphics[width=0.32\columnwidth]{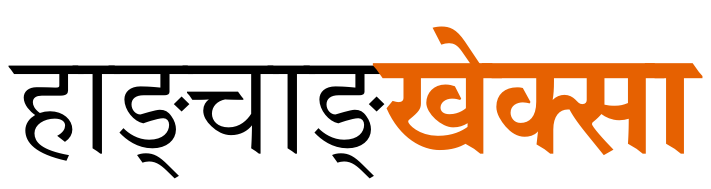}
    \end{tabular}
    \caption{Our model fixes many mixed script and uncommon diacritics errors such as (a) and (b). In rare cases, it ``over-corrects" the first pass OCR transcription, introducing errors such as (c) and (d).}
    \label{fig:error_examples}
\end{figure*}

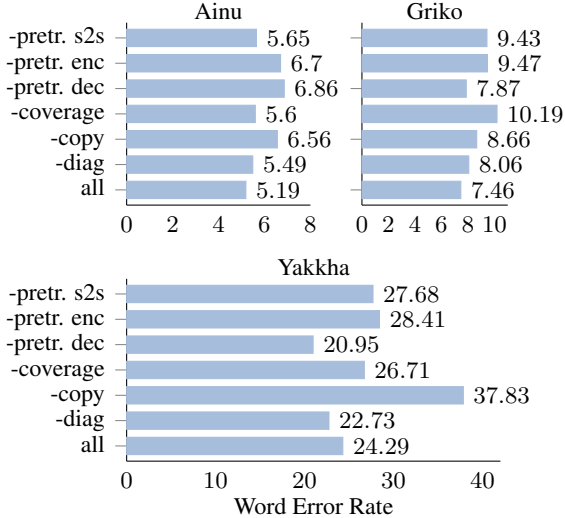
\begin{figure}[t]
    \definecolor{graphblue}{HTML}{A6BDDB}
\pgfplotstableread[row sep=\\,col sep=&]{
det & Ainu & Griko & Yakkha \\
all & 5.19 & 7.46 & 24.29 \\
diag & 5.49 & 8.06 & 22.73 \\
copy & 6.56 & 8.66 & 37.83 \\
coverage & 5.60 & 10.19 & 26.71 \\
dec & 6.86 & 7.87 & 20.95 \\
enc & 6.70 & 9.47 & 28.41 \\
seq2seq & 5.65 & 9.43 & 27.68 \\
}\data
\def\mystrut{\vphantom{hp}}

%\vspace{-1em}
\begin{tikzpicture}[trim left=-1.6cm,trim right=0cm]
    \begin{axis}[
            xbar,
            every axis plot post/.style={/pgf/number format/fixed},
            bar width=.23cm,
            width=4cm,
            height=4cm,
            ymajorgrids=false,
            yminorgrids=false,
            xmajorgrids=false,
            %every axis legend/.code={\let\addlegendentry\relax},
            %legend={Wikipedia Size (in million articles)},
            legend style={draw=none,at={(0.2,0.9)},anchor=west},
            symbolic y coords={all,diag,copy,coverage,dec,enc,seq2seq},
            ytick={all,diag,copy,coverage,dec,enc,seq2seq},
            yticklabels={all,-diag,-copy,-coverage,-pretr. dec,-pretr. enc,-pretr. s2s},
            every y tick label/.append style={font=\small\mystrut},
            every x tick label/.append style={font=\small\mystrut},
            tick pos=left,
            %hide y axis,
            axis x line*=bottom,
            axis y line*=left,
            nodes near coords,
            %nodes near coords align={vertical},
            every node near coord/.append style={font=\small,color=black},
            %nodes near coords style={},
            title={\small Ainu},
            title style={yshift=-0.3cm},
            xmin=0,xmax=8,
            %ylabel shift={-1cm},
            %ylabel near ticks,
            %ylabel={},
            %xlabel near ticks,
            %xlabel={WER},
            enlarge x limits=0.0,
            xtick style={draw=none}
        ]
        %\addplot [style={black,postaction={pattern=north east lines},fill=white,mark=none}] table[x=story,y=base]{\ainudata};
        \addplot [style={graphblue,fill=graphblue,mark=none}] table[x=Ainu,y=det]{\data};
        %\legend{Wikipedia Size (in million articles)}
    \end{axis}
\end{tikzpicture}
\begin{tikzpicture}[trim left=-3cm,trim right=0cm]
    \begin{axis}[
            xbar,
            every axis plot post/.style={/pgf/number format/fixed},
            bar width=.23cm,
            width=3.5cm,
            height=4cm,
            ymajorgrids=false,
            yminorgrids=false,
            xmajorgrids=false,
            %every axis legend/.code={\let\addlegendentry\relax},
            %legend={Wikipedia Size (in million articles)},
            legend style={draw=none,at={(0.2,0.9)},anchor=west},
            symbolic y coords={all,diag,copy,coverage,dec,enc,seq2seq},
            xtick={0,2,4,6,8,10},
            yticklabels={,,,,,,},
            every y tick label/.append style={font=\small\mystrut},
            every x tick label/.append style={font=\small\mystrut},
            %tick pos=none,
            %hide y axis,
            axis x line*=bottom,
            axis y line*=left,
            nodes near coords,
            %nodes near coords align={vertical},
            every node near coord/.append style={font=\small,color=black},
            %nodes near coords style={},
            title={\small Griko},
            title style={yshift=-0.3cm},
            xmin=0,xmax=11,
            %ylabel shift={-1cm},
            %ylabel near ticks,
            %ylabel={},
            %xlabel near ticks,
            %xlabel={WER},
            enlarge x limits=0.0,
            xtick style={draw=none}
        ]
        %\addplot [style={black,postaction={pattern=north east lines},fill=white,mark=none}] table[x=story,y=base]{\ainudata};
        \addplot [style={graphblue,fill=graphblue,mark=none}] table[x=Griko,y=det]{\data};
        %\legend{Wikipedia Size (in million articles)}
    \end{axis}
\end{tikzpicture}

\begin{tikzpicture}[trim left=-1.6cm,trim right=0cm]
    \begin{axis}[
            xbar,
            every axis plot post/.style={/pgf/number format/fixed},
            bar width=.23cm,
            width=6.5cm,
            height=4cm,
            ymajorgrids=false,
            yminorgrids=false,
            xmajorgrids=false,
            %every axis legend/.code={\let\addlegendentry\relax},
            %legend={Wikipedia Size (in million articles)},
            legend style={draw=none,at={(0.2,0.9)},anchor=west},
            symbolic y coords={all,diag,copy,coverage,dec,enc,seq2seq},
            ytick={all,diag,copy,coverage,dec,enc,seq2seq},
            yticklabels={all,-diag,-copy,-coverage,-pretr. dec,-pretr. enc,-pretr. s2s},
            every y tick label/.append style={font=\small\mystrut},
            every x tick label/.append style={font=\small\mystrut},
            tick pos=left,
            %hide y axis,
            axis x line*=bottom,
            axis y line*=left,
            nodes near coords,
            %nodes near coords align={vertical},
            every node near coord/.append style={font=\small,color=black},
            %nodes near coords style={},
            title={\small Yakkha},
            title style={yshift=-0.3cm},
            xmin=0,xmax=42,
            xlabel shift={-0.2cm},
            %ylabel near ticks,
            %ylabel={},
            xlabel near ticks,
            xlabel={\small Word Error Rate},
            enlarge x limits=0.0,
            enlarge y limits=0.1,
            xtick style={draw=none},
            % xtick align=outside
            % ytick style={draw=none}
        ]
        %\addplot [style={black,postaction={pattern=north east lines},fill=white,mark=none}] table[x=story,y=base]{\ainudata};
        \addplot [style={graphblue,fill=graphblue,mark=none}] table[x=Yakkha,y=det]{\data};
        %\legend{Wikipedia Size (in million articles)}
    \end{axis}
\end{tikzpicture}
    \caption{WER with model component ablations on the best model setting in \autoref{tab:cer}. ``all" includes all the adaptations we propose. Each ablation removes a single component from the ``all" model, e.g. ``-pretr.~s2s" removes the seq-to-seq model pretraining.}
    \label{fig:ablation}
\end{figure}

\autoref{tab:cer} shows the performance of the baselines and our proposed method for each language. Overall, our method results in an improved CER and WER over existing methods across all three languages. 

The \textsc{Base} system does not improve the recognition rate over the first pass transcription, apart from a small decrease in the Griko WER. The performance on Yakkha, particularly, is significantly worse than \textsc{Fp-Ocr}: likely because the data size of Yakkha is much smaller than that of Griko and Ainu, and the model is unable to learn a reasonable distribution. However, on adding a copy mechanism to the base model in the \textsc{Copy} system, the performance is notably better for both Griko and Yakkha. This indicates that adaptations to the base model that cater to specific characteristics of the post-correction task can alleviate some of the challenges of learning from small amounts of data.

The single-source and the multi-source variants of our proposed method improve over the baselines, demonstrating that our proposed model adaptations can improve recognition even without translations. We see that using the high-resource translations results in better post-correction performance for Griko and Yakkha, but the single-source model achieves better accuracy for Ainu. We attribute this to two factors: the very low error rate of the first pass transcription for Ainu and the relatively high error rate (based on manual inspection) of the OCR on the Japanese translation. Despite being a high-resource language, OCR is difficult due to the complexity of Japanese characters and low scan quality. The noise resulting from the Japanese OCR errors likely hurts the multi-source model.

\subsection{Ablation Studies}

Next, we study the effect of our proposed adaptations and evaluate their benefit to the performance of each language. \autoref{fig:ablation} shows the word error rate with models that remove one adaptation from the model with all the adaptations (``all").

For Ainu and Griko, removing any single component increases the WER, with the complete (\ba\ba all'') method performing the best. There is little variance in the Ainu ablations, likely due to the high-quality first pass transcription. 

Our proposed adaptations add the most benefit for Yakkha, which has the fewest training data and relatively less accurate first pass OCR. The copy mechanism is crucial for good performance, but removing the decoder pretraining (\ba\ba pretr.~dec'') leads to the best scores among all the ablations. The denoising rules used to create the pseudo-target data for Yakkha are likely not accurate since they are derived from only three paragraphs of annotated data. Consequently, using it to pretrain the decoder leads to a poor language model.

\subsection{Error Analysis}

We systematically inspect all the recognition errors in the output of our post-correction model to determine the sources of improvement with respect to the first pass OCR. We also examine the types of errors introduced by the post-correction process.

We observe a \emph{91\% reduction} in the number of errors due to mixed scripts and a \emph{58\% reduction} in the errors due to uncommon characters and diacritics (as defined in \autoref{sec:analysis}). Examples of these are shown in \autoref{fig:error_examples} (a) and (b): mixed script errors such as the $\bm{\chi}$ character in Griko and the glottal stop \textbf{\textipa{\textglotstop}} in Yakkha are successfully corrected by the model. The model is also able to correct uncommon character errors like \textbf{\d{d}} in Griko and {\raisebox{-2.65pt}{\includegraphics[height=9.5pt]{images/dev_char.pdf}}} in Yakkha.

Examples of errors introduced by the model are shown in \autoref{fig:error_examples} (c) and (d). Example (c) is in Griko, where the model incorrectly adds a diacritic to a character. We attribute this to the fact that the first pass OCR does not recognize diacritics well; hence, the model learns to add diacritics frequently while generating the output. Example (d) is in Yakkha. The model inserts several incorrect characters, and can likely be attributed to the lack of a good language model due to the relatively smaller amount of training data we have in Yakkha.  

\section{Related Work}
Post-correction for OCR is well-studied for high-resource languages. Early approaches include lexical methods and weighted finite-state methods (see \citet{schulz-kuhn-2017-multi} for an overview). Recent work has primarily focused on using neural sequence-to-sequence models. \citet{hamalainen-hengchen-2019-paft} use a BiLSTM encoder-decoder with attention for historical English post-correction. Similar to our base model, \citet{dong-smith-2018-multi} use a multi-source model to combine the first pass OCR from duplicate documents in English. 

There has been little work on lower-resourced languages. \citet{kolak-resnik-2005-ocr} present a probabilistic edit distance based post-correction model applied to Cebuano and Igbo, and \citet{krishna-etal-2018-upcycle} show improvements on Romanized Sanksrit OCR by adding a copy mechanism to a neural sequence-to-sequence model.

Multi-source encoder-decoder models have been used for various tasks including machine translation~\cite{zoph-knight-2016-multi,libovicky-helcl-2017-attention} and morphological inflection~\cite{kann-etal-2017-neural,anastasopoulos-neubig-2019-pushing}. Perhaps most relevant to our work is the multi-source model presented by \citet{anastasopoulos+chiang:interspeech2018}, which uses high-resource translations to improve speech transcription of lower-resourced languages.

Finally, \citet{bustamante-etal-2020-data} construct corpora for four endangered languages from text-based PDFs using rule-based heuristics. Data creation from such unstructured text files is an important research direction, complementing our method of extracting data from scanned images.
\section{Conclusion}
This work presents a first step towards extracting textual data in endangered languages from scanned images of paper books. We create a benchmark dataset with transcribed images in three endangered languages: Ainu, Griko, and Yakkha. We propose an OCR post-correction method that facilitates learning from small amounts of data, which results in a 34\% average relative error reduction in both the character and word recognition rates.

As future work, we plan to investigate the effect of using other available data for the three languages (for example, word lists collected by documentary linguists or the additional Griko folk tales collected by~\citet{anastasopoulos-etal-2018-part}). 

Additionally, it would be valuable to examine whether our method can improve the OCR on high-resource languages, which typically have much better recognition rates in the first pass transcription than the endangered languages in our dataset.

Further, we note our use of the Google Vision OCR system to obtain the first pass OCR for our experiments, primarily because it provides script-specific models as opposed to other general-purpose OCR systems that rely on language-specific models (as discussed in \autoref{sec:analysis}). Future work that focuses on overcoming the challenges of applying language-specific models to endangered language texts would be needed to confirm our method's applicability to post-correcting the first pass transcriptions from different OCR systems.

Lastly, given the annotation effort involved, this paper explores only a small fraction of the endangered language data available in linguistic and general-purpose archives.
Future work will focus on large-scale digitization of scanned documents, aiming to expand our OCR benchmark on as many endangered languages as possible, in the hope of both easing linguistic documentation and preservation efforts and collecting enough data for NLP system development in under-represented languages.

\section*{Acknowledgements}
We thank David Chiang, Walter Scheirer, and William Theisen for initial discussions on the project, the University of Notre Dame Library for the scanned ``Kutune Shirka" Ainu-Japanese book, and Josep Quer for the scanned Griko folk-tales book. We also thank Taylor Berg-Kirkpatrick, Shuyan Zhou, Zi-Yi Dou, Yansen Wang, Zhen Fan, and Deepak Gopinath for feedback on the paper.

This material is based upon work supported in part by the National Science Foundation under Grant No. 1761548. Shruti Rijhwani is supported by a Bloomberg Data Science Ph.D. Fellowship.

\bibliographystyle{acl_natbib}
\bibliography{emnlp2020}

\begin{thebibliography}{32}
\expandafter\ifx\csname natexlab\endcsname\relax\def\natexlab#1{#1}\fi

\bibitem[{Anastasopoulos and
  Chiang(2018)}]{anastasopoulos+chiang:interspeech2018}
Antonios Anastasopoulos and David Chiang. 2018.
\newblock Leveraging translations for speech transcription in low-resource
  settings.
\newblock In \emph{Proc. INTERSPEECH}.

\bibitem[{Anastasopoulos et~al.(2018)Anastasopoulos, Lekakou, Quer, Zimianiti,
  DeBenedetto, and Chiang}]{anastasopoulos-etal-2018-part}
Antonios Anastasopoulos, Marika Lekakou, Josep Quer, Eleni Zimianiti, Justin
  DeBenedetto, and David Chiang. 2018.
\newblock \href {https://www.aclweb.org/anthology/C18-1214} {Part-of-speech
  tagging on an endangered language: a parallel {G}riko-{I}talian resource}.
\newblock In \emph{Proceedings of the 27th International Conference on
  Computational Linguistics}, pages 2529--2539, Santa Fe, New Mexico, USA.
  Association for Computational Linguistics.

\bibitem[{Anastasopoulos and Neubig(2019)}]{anastasopoulos-neubig-2019-pushing}
Antonios Anastasopoulos and Graham Neubig. 2019.
\newblock \href {https://doi.org/10.18653/v1/D19-1091} {Pushing the limits of
  low-resource morphological inflection}.
\newblock In \emph{Proceedings of the 2019 Conference on Empirical Methods in
  Natural Language Processing and the 9th International Joint Conference on
  Natural Language Processing (EMNLP-IJCNLP)}, pages 984--996, Hong Kong,
  China. Association for Computational Linguistics.

\bibitem[{Bahdanau et~al.(2015)Bahdanau, Cho, and Bengio}]{bahdanau2015neural}
Dzmitry Bahdanau, Kyunghyun Cho, and Yoshua Bengio. 2015.
\newblock Neural machine translation by jointly learning to align and
  translate.
\newblock In \emph{3rd International Conference on Learning Representations,
  ICLR 2015}.

\bibitem[{Berg-Kirkpatrick et~al.(2013)Berg-Kirkpatrick, Durrett, and
  Klein}]{berg-kirkpatrick-etal-2013-unsupervised}
Taylor Berg-Kirkpatrick, Greg Durrett, and Dan Klein. 2013.
\newblock \href {https://www.aclweb.org/anthology/P13-1021} {Unsupervised
  transcription of historical documents}.
\newblock In \emph{Proceedings of the 51st Annual Meeting of the Association
  for Computational Linguistics (Volume 1: Long Papers)}, pages 207--217,
  Sofia, Bulgaria. Association for Computational Linguistics.

\bibitem[{Bustamante et~al.(2020)Bustamante, Oncevay, and
  Zariquiey}]{bustamante-etal-2020-data}
Gina Bustamante, Arturo Oncevay, and Roberto Zariquiey. 2020.
\newblock \href {https://www.aclweb.org/anthology/2020.lrec-1.356} {No data to
  crawl? monolingual corpus creation from {PDF} files of truly low-resource
  languages in {P}eru}.
\newblock In \emph{Proceedings of The 12th Language Resources and Evaluation
  Conference}, pages 2914--2923, Marseille, France. European Language Resources
  Association.

\bibitem[{Devos(2019)}]{shangaji-elar}
Maud Devos. 2019.
\newblock Shangaji. a maka or swahili language of mozambique. grammar, texts
  and wordlist.
\newblock \url{https://elar.soas.ac.uk/Collection/MPI1029699}.
\newblock Accessed: 2020-02-02.

\bibitem[{Dong and Smith(2018)}]{dong-smith-2018-multi}
Rui Dong and David Smith. 2018.
\newblock \href {https://doi.org/10.18653/v1/P18-1220} {Multi-input attention
  for unsupervised {OCR} correction}.
\newblock In \emph{Proceedings of the 56th Annual Meeting of the Association
  for Computational Linguistics (Volume 1: Long Papers)}, pages 2363--2372,
  Melbourne, Australia. Association for Computational Linguistics.

\bibitem[{Fujii et~al.(2017)Fujii, Driesen, Baccash, Hurst, and
  Popat}]{fujii2017sequence}
Yasuhisa Fujii, Karel Driesen, Jonathan Baccash, Ash Hurst, and Ashok~C Popat.
  2017.
\newblock Sequence-to-label script identification for multilingual ocr.
\newblock In \emph{2017 14th IAPR International Conference on Document Analysis
  and Recognition (ICDAR)}, volume~1, pages 161--168. IEEE.

\bibitem[{Gu et~al.(2016)Gu, Lu, Li, and Li}]{gu-etal-2016-incorporating}
Jiatao Gu, Zhengdong Lu, Hang Li, and Victor~O.K. Li. 2016.
\newblock \href {https://doi.org/10.18653/v1/P16-1154} {Incorporating copying
  mechanism in sequence-to-sequence learning}.
\newblock In \emph{Proceedings of the 54th Annual Meeting of the Association
  for Computational Linguistics (Volume 1: Long Papers)}, pages 1631--1640,
  Berlin, Germany. Association for Computational Linguistics.

\bibitem[{H{\"a}m{\"a}l{\"a}inen and
  Hengchen(2019)}]{hamalainen-hengchen-2019-paft}
Mika H{\"a}m{\"a}l{\"a}inen and Simon Hengchen. 2019.
\newblock \href {https://doi.org/10.26615/978-954-452-056-4_051} {From the paft
  to the fiiture: a fully automatic {NMT} and word embeddings method for {OCR}
  post-correction}.
\newblock In \emph{Proceedings of the International Conference on Recent
  Advances in Natural Language Processing (RANLP 2019)}, pages 431--436, Varna,
  Bulgaria. INCOMA Ltd.

\bibitem[{Hochreiter and Schmidhuber(1997)}]{hochreiter1997long}
Sepp Hochreiter and J{\"u}rgen Schmidhuber. 1997.
\newblock Long short-term memory.
\newblock \emph{Neural computation}, 9(8):1735--1780.

\bibitem[{Howell(1951)}]{howell1951classification}
Richard~W Howell. 1951.
\newblock The classification and description of ainu folklore.
\newblock \emph{The Journal of American Folklore}, 64(254):361--369.

\bibitem[{Ingle et~al.(2019)Ingle, Fujii, Deselaers, Baccash, and
  Popat}]{ingle2019scalable}
R~Reeve Ingle, Yasuhisa Fujii, Thomas Deselaers, Jonathan Baccash, and Ashok~C
  Popat. 2019.
\newblock A scalable handwritten text recognition system.
\newblock \emph{arXiv preprint arXiv:1904.09150}.

\bibitem[{Joshi et~al.(2020)Joshi, Santy, Budhiraja, Bali, and
  Choudhury}]{joshi2020state}
Pratik Joshi, Sebastin Santy, Amar Budhiraja, Kalika Bali, and Monojit
  Choudhury. 2020.
\newblock \href {https://doi.org/10.18653/v1/2020.acl-main.560} {The state and
  fate of linguistic diversity and inclusion in the {NLP} world}.
\newblock In \emph{Proceedings of the 58th Annual Meeting of the Association
  for Computational Linguistics}, pages 6282--6293, Online. Association for
  Computational Linguistics.

\bibitem[{Kann et~al.(2017)Kann, Cotterell, and
  Sch{\"u}tze}]{kann-etal-2017-neural}
Katharina Kann, Ryan Cotterell, and Hinrich Sch{\"u}tze. 2017.
\newblock \href {https://www.aclweb.org/anthology/E17-1049} {Neural
  multi-source morphological reinflection}.
\newblock In \emph{Proceedings of the 15th Conference of the {E}uropean Chapter
  of the Association for Computational Linguistics: Volume 1, Long Papers},
  pages 514--524, Valencia, Spain. Association for Computational Linguistics.

\bibitem[{Kindaichi(1931)}]{kindaichi1931ainu}
Ky{\=o}suke Kindaichi. 1931.
\newblock \emph{Ainu Jojishi Y{\=u}kara no Kenky{\=u} [Research on Ainu Epic
  Yukar]}.
\newblock T{\=o}ky{\=o}: T{\=o}ky{\=o} Bunko.

\bibitem[{Kolak and Resnik(2005)}]{kolak-resnik-2005-ocr}
Okan Kolak and Philip Resnik. 2005.
\newblock \href {https://www.aclweb.org/anthology/H05-1109} {{OCR}
  post-processing for low density languages}.
\newblock In \emph{Proceedings of Human Language Technology Conference and
  Conference on Empirical Methods in Natural Language Processing}, pages
  867--874, Vancouver, British Columbia, Canada. Association for Computational
  Linguistics.

\bibitem[{Krishna et~al.(2018)Krishna, Majumder, Bhat, and
  Goyal}]{krishna-etal-2018-upcycle}
Amrith Krishna, Bodhisattwa~P. Majumder, Rajesh Bhat, and Pawan Goyal. 2018.
\newblock \href {https://doi.org/10.18653/v1/K18-1034} {Upcycle your {OCR}:
  Reusing {OCR}s for post-{OCR} text correction in {R}omanised {S}anskrit}.
\newblock In \emph{Proceedings of the 22nd Conference on Computational Natural
  Language Learning}, pages 345--355, Brussels, Belgium. Association for
  Computational Linguistics.

\bibitem[{Lamraoui and Langlais(2013)}]{yasa-1336}
Fethi Lamraoui and Philippe Langlais. 2013.
\newblock Yet another fast, robust and open source sentence aligner. time to
  reconsider sentence alignment?
\newblock In \emph{XIV Machine Translation Summit}, Nice, France.

\bibitem[{Libovick{\'y} and Helcl(2017)}]{libovicky-helcl-2017-attention}
Jind{\v{r}}ich Libovick{\'y} and Jind{\v{r}}ich Helcl. 2017.
\newblock \href {https://doi.org/10.18653/v1/P17-2031} {Attention strategies
  for multi-source sequence-to-sequence learning}.
\newblock In \emph{Proceedings of the 55th Annual Meeting of the Association
  for Computational Linguistics (Volume 2: Short Papers)}, pages 196--202,
  Vancouver, Canada. Association for Computational Linguistics.

\bibitem[{Mi et~al.(2016)Mi, Sankaran, Wang, and
  Ittycheriah}]{mi-etal-2016-coverage}
Haitao Mi, Baskaran Sankaran, Zhiguo Wang, and Abe Ittycheriah. 2016.
\newblock \href {https://doi.org/10.18653/v1/D16-1096} {Coverage embedding
  models for neural machine translation}.
\newblock In \emph{Proceedings of the 2016 Conference on Empirical Methods in
  Natural Language Processing}, pages 955--960, Austin, Texas. Association for
  Computational Linguistics.

\bibitem[{Neubig et~al.(2017)Neubig, Dyer, Goldberg, Matthews, Ammar,
  Anastasopoulos, Ballesteros, Chiang, Clothiaux, Cohn, Duh, Faruqui, Gan,
  Garrette, Ji, Kong, Kuncoro, Kumar, Malaviya, Michel, Oda, Richardson,
  Saphra, Swayamdipta, and Yin}]{dynet}
Graham Neubig, Chris Dyer, Yoav Goldberg, Austin Matthews, Waleed Ammar,
  Antonios Anastasopoulos, Miguel Ballesteros, David Chiang, Daniel Clothiaux,
  Trevor Cohn, Kevin Duh, Manaal Faruqui, Cynthia Gan, Dan Garrette, Yangfeng
  Ji, Lingpeng Kong, Adhiguna Kuncoro, Gaurav Kumar, Chaitanya Malaviya, Paul
  Michel, Yusuke Oda, Matthew Richardson, Naomi Saphra, Swabha Swayamdipta, and
  Pengcheng Yin. 2017.
\newblock Dynet: The dynamic neural network toolkit.
\newblock \emph{arXiv preprint arXiv:1701.03980}.

\bibitem[{{Rigaud} et~al.(2019){Rigaud}, {Doucet}, {Coustaty}, and
  {Moreux}}]{8978127}
C.~{Rigaud}, A.~{Doucet}, M.~{Coustaty}, and J.~{Moreux}. 2019.
\newblock {ICDAR} 2019 competition on post-{OCR} text correction.
\newblock In \emph{2019 International Conference on Document Analysis and
  Recognition (ICDAR)}, pages 1588--1593.

\bibitem[{Schackow(2012)}]{yakkha-elar}
Diana Schackow. 2012.
\newblock Documentation and grammatical description of yakkha, nepal.
\newblock \url{https://elar.soas.ac.uk/Collection/MPI186180}.
\newblock Accessed: 2020-02-02.

\bibitem[{Schackow(2015)}]{Schackow_2015}
Diana Schackow. 2015.
\newblock \href {https://doi.org/10.26530/oapen_603340} {\emph{A grammar of
  Yakkha}}.
\newblock Language Science Press.

\bibitem[{Schnober et~al.(2016)Schnober, Eger, Do~Dinh, and
  Gurevych}]{schnober-etal-2016-still}
Carsten Schnober, Steffen Eger, Erik-L{\^a}n Do~Dinh, and Iryna Gurevych. 2016.
\newblock \href {https://www.aclweb.org/anthology/C16-1160} {Still not there?
  comparing traditional sequence-to-sequence models to encoder-decoder neural
  networks on monotone string translation tasks}.
\newblock In \emph{Proceedings of {COLING} 2016, the 26th International
  Conference on Computational Linguistics: Technical Papers}, pages 1703--1714,
  Osaka, Japan. The COLING 2016 Organizing Committee.

\bibitem[{Schulz and Kuhn(2017)}]{schulz-kuhn-2017-multi}
Sarah Schulz and Jonas Kuhn. 2017.
\newblock \href {https://doi.org/10.18653/v1/D17-1288} {Multi-modular
  domain-tailored {OCR} post-correction}.
\newblock In \emph{Proceedings of the 2017 Conference on Empirical Methods in
  Natural Language Processing}, pages 2716--2726, Copenhagen, Denmark.
  Association for Computational Linguistics.

\bibitem[{See et~al.(2017)See, Liu, and Manning}]{see-etal-2017-get}
Abigail See, Peter~J. Liu, and Christopher~D. Manning. 2017.
\newblock \href {https://doi.org/10.18653/v1/P17-1099} {Get to the point:
  Summarization with pointer-generator networks}.
\newblock In \emph{Proceedings of the 55th Annual Meeting of the Association
  for Computational Linguistics (Volume 1: Long Papers)}, pages 1073--1083,
  Vancouver, Canada. Association for Computational Linguistics.

\bibitem[{Stomeo(1980)}]{stomeo1980racconti}
Paolo Stomeo. 1980.
\newblock \emph{Racconti greci inediti di {S}ternat\'{i}a}.
\newblock La nuova Ellade, s.I.

\bibitem[{Tu et~al.(2016)Tu, Lu, Liu, Liu, and Li}]{tu-etal-2016-modeling}
Zhaopeng Tu, Zhengdong Lu, Yang Liu, Xiaohua Liu, and Hang Li. 2016.
\newblock \href {https://doi.org/10.18653/v1/P16-1008} {Modeling coverage for
  neural machine translation}.
\newblock In \emph{Proceedings of the 54th Annual Meeting of the Association
  for Computational Linguistics (Volume 1: Long Papers)}, pages 76--85, Berlin,
  Germany. Association for Computational Linguistics.

\bibitem[{Zoph and Knight(2016)}]{zoph-knight-2016-multi}
Barret Zoph and Kevin Knight. 2016.
\newblock \href {https://doi.org/10.18653/v1/N16-1004} {Multi-source neural
  translation}.
\newblock In \emph{Proceedings of the 2016 Conference of the North {A}merican
  Chapter of the Association for Computational Linguistics: Human Language
  Technologies}, pages 30--34, San Diego, California. Association for
  Computational Linguistics.

\end{thebibliography}

\newpage
\appendix
\section{Appendix}
\label{sec:appendix}
\subsection{Implementation Details}

\noindent
The hyperparameters used are:
\begin{itemize}[itemsep=0pt]
    \item Character embedding size = 128
    \item Number of LSTM layers = 1
    \item Hidden state size of the LSTM = 256
    \item Attention size = 256
    \item Beam size = 4
    \item For the diagonal loss, $j$ = 3
    \item Minibatch size for training = 1
    \item Maximum number of epochs = 150
    \item Patience for early stopping = 10 epochs
    \item Pretraining epochs for encoder/decoder = 10
    \item Pretraining epochs for seq-to-seq model = 5
\end{itemize}

\noindent
We use the same values of the hyperparameters for each language and all the systems. We select the best model with early stopping on the character error rate of the validation set.

\subsection{Additional Experimental Results}

\begin{table}[b]
    \centering
    \begin{tabular}{@{}lrr@{}}
    \toprule
    Model & CER & WER \\
    \midrule
    \textsc{Fp-Ocr} & $8.90$ & $31.64$ \\
    \textsc{Base} & $70.89$ & $100.00$ \\
    \textsc{Copy} & $11.60$ & $26.74$ \\
    \textsc{Ours} & $7.95$ & $20.83$ \\
    \bottomrule
    \end{tabular}
    \caption{Character error rate (CER) and word error rate (WER) for the Yakkha dataset with the multisource model that uses the OCR on Nepali as the high-resource translation. The table shows the mean over five random runs.}
    \label{tab:nepali}
    \vspace{2em}
\end{table}

\paragraph{Performance on Yakkha with Nepali}\autoref{tab:nepali} shows the performance for the Yakkha dataset when using Nepali as the high-resource translation input to the multisource model. The performance is similar to those of the experiments using the English translations, as presented in \autoref{tab:cer}.

\paragraph{Standard deviation on the main results}\autoref{tab:stddev_cer} and \autoref{tab:stddev_wer} show the character error rate and word error rate respectively including the standard deviation over five randomly seeded runs, corresponding to the systems described in \autoref{tab:cer}. 

\begin{table}[tb]
    \centering
    \small
    (a) Ainu\\
    \begin{tabular}{l|r@{\ \ \ }r}
    \toprule
        Model & \multicolumn{1}{c}{Multi} & \multicolumn{1}{c}{Single} \\
        \midrule
        \textsc{Fp-Ocr} & \multicolumn{1}{c}{--} & $1.34$ \\
        \textsc{Base} & $1.56 \pm 0.23$ & $1.41 \pm 0.16$ \\
        \textsc{Copy} & $2.04 \pm 0.62$ & $1.99 \pm 0.41$ \\
        \textsc{Ours} & $0.92 \pm 0.05$ & $\boldsymbol{0.80} \pm 0.07$ \\
    \bottomrule
    \end{tabular}\\
    \vspace{1em}
    (b) Griko\\
    \begin{tabular}{l|r@{\ \ \ }r}
    \toprule
        Model & \multicolumn{1}{c}{Multi} & \multicolumn{1}{c}{Single} \\
        \midrule
        \textsc{Fp-Ocr} & \multicolumn{1}{c}{--} & $3.27$ \\
        \textsc{Base} & $6.78 \pm 0.62$ & $5.95 \pm 0.52$ \\
        \textsc{Copy} & $2.54 \pm 0.31$ & $2.28 \pm 0.28$ \\
        \textsc{Ours} & $\boldsymbol{1.66} \pm 0.03$ & $1.70 \pm 0.21$ \\
    \bottomrule
    \end{tabular}\\
    \vspace{1em}
    (c) Yakkha\\
    \begin{tabular}{l|r@{\ \ \ }r}
    \toprule
        Model & \multicolumn{1}{c}{Multi} & \multicolumn{1}{c}{Single} \\
        \midrule
        \textsc{Fp-Ocr} & \multicolumn{1}{c}{--} & $8.90$ \\
        \textsc{Base} & $70.39 \pm 0.49$ & $71.71 \pm 0.71$  \\
        \textsc{Copy} & $14.77 \pm 0.97$ & $12.30 \pm 2.39$ \\
        \textsc{Ours} & $\boldsymbol{7.75} \pm 0.46$ & $8.44 \pm 0.90$ \\
    \bottomrule
    \end{tabular}
    \caption{Mean and standard deviation of the character error rate with 10-fold cross-validation over five random seeds. The results presented are the same as \autoref{tab:cer} with the added information of standard deviation. The best models for each language are \textbf{highlighted}.}
    \label{tab:stddev_cer}
\end{table}

\begin{table}[tb]
    \centering
    \small
    (a) Ainu\\
    \begin{tabular}{l|r@{\ \ \ }r}
    \toprule
        Model & \multicolumn{1}{c}{Multi} & \multicolumn{1}{c}{Single} \\
        \midrule
        \textsc{Fp-Ocr} & \multicolumn{1}{c}{--} & $6.27$ \\
        \textsc{Base} & $8.56 \pm 1.01$ & $7.88 \pm 0.64$   \\
        \textsc{Copy} & $9.48 \pm 3.07$ & $8.57 \pm 1.45$ \\
        \textsc{Ours} & $5.75 \pm 0.24$ & $\boldsymbol{5.19} \pm 0.31$ \\
    \bottomrule
    \end{tabular}\\
    \vspace{1em}
    (b) Griko\\
    \begin{tabular}{l|r@{\ \ \ }r}
    \toprule
        Model & \multicolumn{1}{c}{Multi} & \multicolumn{1}{c}{Single} \\
        \midrule
        \textsc{Fp-Ocr} & \multicolumn{1}{c}{--} & $15.63$ \\
        \textsc{Base} & $15.13 \pm 0.99$ & $13.67 \pm 1.17$ \\
        \textsc{Copy} & $9.33 \pm 0.49$ & $8.90 \pm 0.51$ \\
        \textsc{Ours} & $\boldsymbol{7.46} \pm 0.09$ & $7.51 \pm 0.31$ \\
    \bottomrule
    \end{tabular}\\
    \vspace{1em}
    (c) Yakkha\\
    \begin{tabular}{l|r@{\ \ \ }r}
    \toprule
        Model & \multicolumn{1}{c}{Multi} & \multicolumn{1}{c}{Single} \\
        \midrule
        \textsc{Fp-Ocr} & \multicolumn{1}{c}{--} & $31.64$ \\
        \textsc{Base} & $98.15 \pm 1.55$ & $99.10 \pm 2.20$  \\
        \textsc{Copy} & $30.36 \pm 1.39$ & $27.81 \pm 1.65$ \\
        \textsc{Ours} & $\boldsymbol{20.95} \pm 1.04$ & $21.33 \pm 0.53$ \\
    \bottomrule
    \end{tabular}
    \caption{Mean and standard deviation of the word error rate with 10-fold cross-validation over five random seeds. The results presented are the same as \autoref{tab:cer} with the added information of standard deviation. The best models for each language are \textbf{highlighted}.}
    \label{tab:stddev_wer}
    \vspace{1em}
\end{table}

\end{document}